\documentclass[11pt,a4paper]{article}
\usepackage[hyperref]{acl2020}
\usepackage{times}
\usepackage{latexsym}

\usepackage{microtype}

\usepackage{tabularx}
\usepackage{multirow}
\usepackage{CJKutf8}
\usepackage[vietnamese,english]{babel}

\aclfinalcopy % Uncomment this line for the final submission
\title{VBD-MT Chinese$\leftrightarrow$Vietnamese Translation Systems for VLSP 2022}

\author{Hai Long Trieu, Song Kiet Bui, Tan Minh Tran, Van Khanh Tran, Hai An Nguyen \\
  Department of Natural Language Processing\\
  Virtual Assistant Technology Center\\
  VinBigData, Hanoi, Vietnam\\
  \texttt{\{v.longth12,v.kietbs,v.minhtt32,v.khanhtv13,v.annh24\}@vinbigdata.com} 
  }

\date{}

\begin{document}
\maketitle
\begin{abstract}
We present our systems participated in the VLSP 2022 machine translation shared task. In the shared task this year, we participated in both translation tasks, i.e., Chinese-Vietnamese and Vietnamese-Chinese translations. We build our systems based on the neural-based Transformer model with the powerful multilingual denoising pre-trained model mBART. The systems are enhanced by a sampling method for backtranslation, which leverage large scale available monolingual data. Additionally, several other methods are applied to improve the translation quality including ensembling and post-processing. We achieve 38.9 BLEU on Chinese-Vietnamese and 38.0 BLEU on Vietnamese-Chinese on the public test sets, which outperform several strong baselines.
\end{abstract}

\section{Introduction}
\label{sec:intro}
For the VLSP machine translation shared task this year (VLSP 2022),\footnote{\url{https://vlsp.org.vn/vlsp2022/eval/mt}} there are two translation tasks, i.e., Chinese-Vietnamese and Vietnamese-Chinese translations. This translation task between these two languages is challenging due to limited bilingual training data although the task has been investigated in several previous work~\cite{van2022kc4mt,tran2016character,zhao2013vietnamese}. There are few powerful translation systems and available large-scale bilingual corpora for this language pair, which cause a gap for machine translation research.

In this work, we aim at investigating and building a strong machine translation (MT) system between Chinese and Vietnamese. For this goal, we first review and compare several strong baseline systems on this language pair. We then leverage monolingual data to enhance the low-resource bilingual data since the provided bilingual training data is very limited (300K bilingual sentences). We also investigate the translation output more deeply and apply post-processing approaches to quickly tackle several typical errors such as translating datetime or numeric values. We report experimental results on the validation and public test data, which show the performance of our baseline systems as well as improvement gains from our approaches. Several analyses are also presented including test samples.

\section{Data}
\label{sec:data}

\subsection{Bilingual Data}
\label{subsec:bidata}
For bilingual data, there are 300K (thousands) parallel sentences provided by the shared task. This can be seen as a low-resource machine translation task in comparison with large-scale bilingual training data, which contain millions of pallel sentences such as Czech-English, English-Chinese, French-German ~\cite{barrault-etal-2020-findings,akhbardeh-etal-2021-findings}. The limited data causes the task challenging but interesting to investigate appropriate approaches and models.

\subsection{Monolingual Data}
\label{subsec:monodata}
In addition to the bilingual data, the shared task also provided large scale monolingual datasets including 19M (millions) sentences for Chinese and 25M sentences for Vietnamese, which can be utilized as an additional resource to enhance translation models.

The monolingual datasets are used for backtranslation (Section~\ref{subsec:backtranslation}). By analyzing training and validation datasets, we found that most sentences are in the length of 10 to 60 words (cover more than 90\%). Therefore, we first quickly apply a filtering method by sentence length, where keeping only sentences in the sentence length range of 10-60. By doing this, we obtained 6M sentences. Due to time limitation, we do not make use the whole but a subset of the data. We randomly selected 1.5M sentences from the filtered 6M data for backtranslation.

\begin{table}[t!]
\centering
\begin{tabular}{l|r|r|r}
\hline \textbf{Data} & \textbf{\#Sents} & \textbf{\#Vocab} & \textbf{Avg.Len} \\ \hline
BiData \\
train (zh) & 300K & 8.3K & 23 \\
train (vi) & 300K & 90.7K & 19\\
\hline
valid (zh) & 1K & 1.6K & 29 \\
valid (vi) & 1K & 3.6K & 21  \\
\hline
test (zh) & 1K & 1.6K & 42 \\
test (vi) & 1K & 3.7K & 36 \\
\hline
MonoData (zh) & 19.0M & 201K & 80 \\
MonoData (vi) & 25.3M & 3,995K & 37 \\
\hline
\end{tabular}
\caption{\label{tab:dataset} Statistics of bilingual and monolingual datasets (with the numbers of sentences (\#Sents), vocabulary size (\#Vocab) and average sentence length (Avg.Len)) of Chinese (zh) and Vietnamese (vi).}
\end{table}

\subsection{Evaluation Data}
For evaluation, the validation set and public test sets contain 1K sentences. We used the officially provided validation set for tuning hyper-parameters and saving model checkpoints, and the systems are compared based on the public test sets.

We present the statistics of these datasets in Table~\ref{tab:dataset}. \footnote{The vocabulary size is an estimation, when we calculate each Chinese character as a token, when Vietnamese tokens are split by white spaces.}

\section{System Overview}
\label{sec:system}

\subsection{Baseline}
We build our baseline systems based on the well-known and powerful Transformer model~\cite{vaswani2017attention}. Our systems are based on the \textit{Fairseq}~\cite{ott2019fairseq} PyTorch implementation, which has achieved state-of-the-art (SOTA) performance on competitions~\cite{tran2021facebook,akhbardeh-etal-2021-findings}. This is confirmed by our preliminary experiments when we compared several systems to select a strong baseline (we present the detail results in Section~\ref{subsec:baseline-selection}). The model is optimized using the Adam optimizer~\cite{kingma2014adam}. We used learning rate $3e-5$, dropout $0.3$, trained with early stopping, and saved models after every $5,000$ steps. We trained our models on 1 NVIDIA GeForce 11GB GPU. We present detail hyper-parameters in Appendix~\ref{appendix:hyper-params}.

The model is finetuned on the mBART~\cite{liu2020multilingual}, which is trained on 25 languages (\textit{mBART-25})\footnote{We also tried the \textit{mBART-50}~\cite{tang2020multilingual}, which is trained on 50 languages, but our preliminary experiments showed that the \textit{mBART-25} is better.} including Chinese and Vietnamese.

We tokenized texts by the well-known \textit{SentencePiece}~\cite{sennrich2016neural}. For evaluation metric, we used the commonly used \textit{SacreBLEU}~\cite{post2018call}.

\begin{table}[t!]
\centering
\begin{tabular}{l|r|r|r}
\hline \textbf{Data} & \textbf{\#Sents} & \textbf{\#Vocab} & \textbf{Avg.Len} \\ \hline
Mono (zh) & 1.5M & 10K & 54 \\
Mono (vi) & 1.5M & 357K & 29 \\
\hline
Synthetic (zh-vi) &  \\
train (zh) & 211K & 17K & 27 \\
train (vi) & 211K & 22K & 29 \\
\hline
Synthetic (vi-zh)  \\ 
train (vi) & 403K & 21K & 45 \\
train (zh) & 403K & 35K & 41 \\
\hline
\end{tabular}
\caption{\label{tab:mono-data} Synthetic data from backtranslation (Mono: the monolingual data used for backtranslation after filtering by sentence length; Synthetic: the synthetic data generated by backtranslation)}
\end{table}

\subsection{Pre-trained model pruning}
We used the \textit{mBART-25}~\cite{liu2020multilingual} for finetuning the baseline systems. The vocabulary size is 250K, which caused the out-of-memory issue when we trained on our NVIDIA GeForce 11GB GPU. We therefore filtered the vocabulary, which contains only the vocabulary of the provided bilingual datasets and the 6M filtered monolingual datasets. Only the embeddings correspond to the filtered vocabulary are kept. As a result, the filtered vocabulary size becomes 67K, which is approximately four times smaller than the original vocabulary, and we can train our systems without a more powerful server. This technique is a minor step, but it may be useful for training systems on limited computational resources.

\subsection{Backtranslation}
\label{subsec:backtranslation}

Leveraging available large scale monolingual data via backtranslation to improve MT systems is a common strategy and has been utilized in previous work \cite{sennrich2016improving,tran2021facebook,akhbardeh-etal-2021-findings}. In this work, we utilize backtranslation using the top-$k$ sampling method~\cite{edunov2018understanding} to generate synthetic data, which select the top-$k$ highest scoring outputs ($k=5$) at every time step. The method has shown to provide richer training signal, which is better than the maximum a posteriori (MAP) beam search, which reduces the diversity and richness of the generated source translations.

We used the baseline systems trained on the provided bilingual data to translate the selected 1.5M monolingual data. As a result, we obtained 211K (zh-vi) and 403K (vi-zh) sentence pairs for back-translated datasets. Table~\ref{tab:mono-data} describes the back-translated data.

\subsection{Ensembling}
Ensembling weights is also a strategy to improve machine learning models~\cite{shahhosseini2022optimizing} including MT systems~\cite{tran2021facebook}. In our systems, we save different epoch checkpoints and calculate the ensembled model by calculating the average of models' weights of the last $N$ checkpoints. We tried different values of $N$ and found that $N=5$ is the best value for our systems.

\subsection{Post-Processing}

Translating numbers look straightforward but has been shown as a challenging task~\cite{wang2021easy}. When we conducted analyses on translation output, we found that specific data types such as date-time and numeric values can be mis-translated, which is a big problem when translating the wrong values of human or currency. The task can be solved by combining with named entity task~\cite{mota2022fast}, but we leave such method for another extended work. In the scope of this work, we simply create a set of patterns for post-processing to edit the translations of date-time and numeric values. For instance, we illustrate several patterns such as the followings.
\begin{itemize}
    \item \begin{CJK*}{UTF8}{gbsn}\textbf{XY亿}\end{CJK*} $\rightarrow$ \textbf{XY} $* 100$ millions
    \item \begin{CJK*}{UTF8}{gbsn}\textbf{XY万}\end{CJK*} $\rightarrow$ \textbf{XY} $* 10$ thousands
\end{itemize}

The patterns are simple yet effective. We discuss the results with corrected samples in Section~\ref{subsec:post-process-result}.

\begin{table}[t!]
\centering
\begin{tabular}{l|cc}
\hline \textbf{Model} & \textbf{Valid} & \textbf{Test} \\ \hline
Fairseq~\cite{ott2019fairseq} & 34.8 & \textbf{38.0} \\
\hline
Transformer & \\
+ RAML~\cite{norouzi2016reward} & 35.3 & 37.0 \\
+ Point.Gen~\cite{enarvi2020generating} & 34.4 & N/A \\
+ Bi-Simcut~\cite{gao-etal-2022-bi} & 33.1 & N/A \\
\hline
MSP~\cite{tan2022msp} & 29.4 & N/A \\
\hline
\end{tabular}
\caption{\label{tab:zh2vi-baselines-results} Compare baselines on Chinese-Vietnamese translation ((N/A): several results are yet completed due to time limitation). Best scores (BLEU) are bold.}
\end{table}

\section{Experiments and Results}

\subsection{Baseline selection}
\label{subsec:baseline-selection}

In order to investigate and build a strong baseline for this task, we first review and compare several existing and recent proposed models. In particular, we compared the following systems.
\begin{itemize}
    \item \textbf{Fairseq}~\cite{ott2019fairseq}: a PyTorch implementation, which has shown the SOTA performance on competitions including the well-known WMT shared tasks~\cite{barrault-etal-2020-findings,akhbardeh-etal-2021-findings}.
    \item \textbf{RAML}~\cite{norouzi2016reward}: used a reward augmented maximum likelihood, which tries to optimize task reward (loss) used for test evaluation, and shown the contribution for neural sequence to sequence models including machine translation.
    \item \textbf{Point.Gen}~\cite{enarvi2020generating}: uses a pointer-generator network to facilitate the same parts of a source sentence (such as person names) to a target sentence.
    \item \textbf{Bi-Simcut}\cite{gao-etal-2022-bi}: a training strategy using a regularization method to forces the consistency between the output distributions of the original and the cutoff sentence pairs to boost translation performance.     
    \item  \textbf{Multi-Stage Prompting (MSP)}~\cite{tan2022msp}: a method uses different continuous prompts for shifting from pre-trained models to translation tasks better.
\end{itemize}

It is noted that for RAML, Bi-Simcut, and Pointer Generator, we used the Huggingface's implementation.\footnote{\url{https://github.com/huggingface/transformers}}

The compared results are presented in Table~\ref{tab:zh2vi-baselines-results}. From these preliminary experiments, we found that the Fairseq obtained the SOTA performance on the public test set. Therefore, we selected the Fairseq implementation for our baseline system.

\subsection{Our systems' results}
% The results of our submitted systems on the validation and public test sets are presented in Tables~\ref{tab:zh2vi-results} and \ref{tab:vi2zh-results}.

\begin{table}[t!]
\centering
\begin{tabular}{l|cc}
\hline \textbf{Model} & \textbf{Valid} & \textbf{Test} \\ \hline
Baseline & 34.8 & 38.0 \\
\hline
Backtranslation (B) & 34.8 & 38.8 \\
(B)+Ensembling (E) & 34.8 & \textbf{38.9} \\
(B)+(E)+Post-Processing & 34.8 & \textbf{38.9} \\
\hline
\end{tabular}
\caption{\label{tab:zh2vi-results} Results on Chinese-Vietnamese translation}
\end{table}

\paragraph{Chinese-Vietnamese translation} The results of Chinese-Vietnamese are presented in Table~\ref{tab:zh2vi-results}. The baseline model gains 38.0 BLEU, which is relatively high when comparing with recent work for this language pair~\cite{van2022kc4mt}. We achieve 38.8 BLEU when leveraging backtranslation, which improves +0.8 BLEU point. Ensembling gains +0.1 BLEU point improvement, while post-processing keeps the same BLEU performance.

\paragraph{Vietnamese-Chinese translation} For Vietnamese-Chinese task, we presented the results in Table~\ref{tab:vi2zh-results}. We also obtain a relatively high performance with the baseline (37.8 BLEU point), which is somehow equivalent with the Chinese-Vietnamese task. Backtranslation improves +0.2 BLEU point, while ensembling and post-processing keep the same performance.

For our submitted systems for the shared task, we used the best performance systems, which are the combination of the baseline, backtranslation, ensembling, and post-processing to produce the output. Although the post-processing step does not improve the performance in BLEU scores, we found that we still obtain better and correct translation output when we analyzed translated samples. We discuss the post-processing output in the next Section.

\subsection{Post-processing}
\label{subsec:post-process-result}
We conducted a post-processing step to correct the translation output, which we focus on correcting numeric and date-time values. We manually check the output of the test set and found that mis-translated output can be can be edited correctly. For instance, the number \begin{CJK*}{UTF8}{gbsn}\textbf{400亿}\end{CJK*} (40 billion) and date \begin{CJK*}{UTF8}{gbsn}\textbf{2021年12月1日}\end{CJK*} (December 1st, 2021) (as presented in Table~\ref{tab:post-process-samples}) are correctly revised via the post-processing step.

\subsection{Human evaluation}
Besides the systems' BLEU performance, we would like to analyze the translation quality from the expert viewpoint. Therefore, we conducted human evaluation to investigate the actual translation output. It is noted that this is not human evaluation conducted by the shared task organizers, but we invited a native speaker, who is fluent in both Chinese and Vietnamese to check and analyze a small set of the translated sentences.

\begin{table}[t!]
\centering
\begin{tabular}{l|cc}
\hline \textbf{Model} & \textbf{Valid} & \textbf{Test} \\ \hline
Baseline & 32.2 & 37.8 \\
\hline
Backtranslation (B) & 32.9 & 38.0 \\
(B)+Ensembling (E) & 32.9 & \textbf{38.0} \\
(B)+(E)+Post-Processing & 32.9 & \textbf{38.0} \\
\hline
\end{tabular}
\caption{\label{tab:vi2zh-results} Results on Vietnamese-Chinese translation}
\end{table}

In particular, for each translation task, we randomly selected a small set (20 translated samples) for the human expert to evaluate and analyze. For each sample, there are two outputs, which are from the baseline and the submitted systems, and we shuffled the outputs so that the expert does not know an output comes from which system. We let the expert to assign a score for each output (scoring $1-10$ point), and give comments for each output. From this evaluation, we gain several following observations.
\begin{itemize}
    \item The submitted systems produce better translations than the baseline systems (received higher scores) in most cases (more than $70\%$ of the samples), which confirm the improvement shown in BLEU scores.
    \item The meaning of current translated output is acceptable to some extent when most cases receive the scores from $7$ to $10$ ($80\%$ of the samples).
    \item Common issues in translations still remain such as: incorrect translations of person names; only correctly translate a part of content or missing main content, unclear translations; using inappropriate translated words, etc.
\end{itemize}

\begin{table*}[t!]
\centering
\begin{tabularx}{\textwidth}{l|X}
\hline \textbf{Data} & \textbf{Sample} \\ \hline
Input (zh) & 
\begin{CJK*}{UTF8}{gbsn} 原因在于澳大利亚决定取消总额\textcolor{green}{400亿}美元的向法国采购核潜艇合同，转而与美国和英国开展联合项目。\end{CJK*}
\\ \hline
Translated (vi) & \begin{otherlanguage}{vietnamese} Nguyên nhân là do Australia quyết định hủy hợp đồng mua tàu ngầm hạt nhân trị giá \textcolor{red}{4 tỷ} USD cho Pháp, chuyển sang triển khai dự án chung với Mỹ và Anh. \end{otherlanguage}
\\ \hline
Post-processed (vi) & \begin{otherlanguage}{vietnamese} Nguyên nhân là do Australia quyết định hủy hợp đồng mua tàu ngầm hạt nhân trị giá \textcolor{blue}{40 tỷ} USD cho Pháp, chuyển sang triển khai dự án chung với Mỹ và Anh. \end{otherlanguage} \\
\hline
Meaning (English) & The reason is that Australia decided to cancel the \textcolor{blue}{\$ 40 billion} contract to buy nuclear submarines for France, moving to implement a joint project with the US and UK. \\
\hline
\hline
Input (zh) & 
\begin{CJK*}{UTF8}{gbsn} 目前，美国和欧盟都期待在\textcolor{green}{2021年12月1日}前达成解决钢铁和铝贸易争端的协议。\end{CJK*}
\\ \hline
Translated (vi) & \begin{otherlanguage}{vietnamese} Hiện cả Mỹ và EU đều trông đợi một thỏa thuận giải quyết tranh chấp thương mại thép và nhôm trước ngày \textcolor{red}{1/1/2021}. \end{otherlanguage}
\\ \hline
Post-processed (vi) & \begin{otherlanguage}{vietnamese} Hiện cả Mỹ và EU đều trông đợi một thỏa thuận giải quyết tranh chấp thương mại thép và nhôm trước ngày \textcolor{blue}{1/12/2021}. \end{otherlanguage} \\
\hline
Meaning (English) & Currently, both the US and EU expect an agreement to settle the steel and aluminum trade dispute before \textcolor{blue}{December 1, 2021}. \\
\hline
\end{tabularx}
\caption{\label{tab:post-process-samples} Samples of number and date (green) are mis-translated (red) and corrected by post-processing (blue)}
\end{table*}

\subsection{Limitations}
Though we achieve promising results with the baseline systems and improvement from our approaches, there are still several limitations. First, experiments on comparing systems to choose a strong baseline are yet fully completed. Second, we only leveraged a subset of the monolingual data (about 10\% or less of the provided data), which has not completely utilized the large scale available data. Third, the performance may be better if we use the full vocabulary of the pretrained \textit{mBART} model instead of pruning the vocabulary, although we are able to get benefits from this pruning when powerful servers are unavailable. Fourth, though we achieve several interesting and useful analyses from the human evaluation, it is better to conduct the evaluation on a larger number of samples.

\section{Conclusion}
In this paper, we describe our systems participated in the VLSP 2022 machine translation shared task for the Chinese-Vietnamese language pair. Our neural-based systems are built based on the Transformer model using the \textit{Fairseq} framework. The model is finetuned on the \textit{mBART} pre-trained model on 25 languages. The model is enhanced by leveraging monolingual data using an upsampling method. Additionally, we create a post-processing step to correct mis-translated output, which we focus on numeric data such as numbers and date-time values. Furthermore, models are ensembled based on weight averaging. We conducted various experiments to select a strong baseline and compare different training settings and approaches. We achieve a relatively high baseline performance and improvement from our approaches, which are confirmed by both empirical experiments (BLEU) and human evaluation. We also present and discuss analyses on translation output as well as point out several limitations, which should be solved to improve the systems in future research.

\bibliography{vlsp22}

\begin{thebibliography}{22}
\expandafter\ifx\csname natexlab\endcsname\relax\def\natexlab#1{#1}\fi

\bibitem[{Akhbardeh et~al.(2021)Akhbardeh, Arkhangorodsky, Biesialska, Bojar,
  Chatterjee, Chaudhary, Costa-jussa, Espa{\~n}a-Bonet, Fan, Federmann,
  Freitag, Graham, Grundkiewicz, Haddow, Harter, Heafield, Homan, Huck,
  Amponsah-Kaakyire, Kasai, Khashabi, Knight, Kocmi, Koehn, Lourie, Monz,
  Morishita, Nagata, Nagesh, Nakazawa, Negri, Pal, Tapo, Turchi, Vydrin, and
  Zampieri}]{akhbardeh-etal-2021-findings}
Farhad Akhbardeh, Arkady Arkhangorodsky, Magdalena Biesialska, Ond{\v{r}}ej
  Bojar, Rajen Chatterjee, Vishrav Chaudhary, Marta~R. Costa-jussa, Cristina
  Espa{\~n}a-Bonet, Angela Fan, Christian Federmann, Markus Freitag, Yvette
  Graham, Roman Grundkiewicz, Barry Haddow, Leonie Harter, Kenneth Heafield,
  Christopher Homan, Matthias Huck, Kwabena Amponsah-Kaakyire, Jungo Kasai,
  Daniel Khashabi, Kevin Knight, Tom Kocmi, Philipp Koehn, Nicholas Lourie,
  Christof Monz, Makoto Morishita, Masaaki Nagata, Ajay Nagesh, Toshiaki
  Nakazawa, Matteo Negri, Santanu Pal, Allahsera~Auguste Tapo, Marco Turchi,
  Valentin Vydrin, and Marcos Zampieri. 2021.
\newblock \href {https://aclanthology.org/2021.wmt-1.1} {Findings of the 2021
  conference on machine translation ({WMT}21)}.
\newblock In \emph{Proceedings of the Sixth Conference on Machine Translation},
  pages 1--88, Online. Association for Computational Linguistics.

\bibitem[{Barrault et~al.(2020)Barrault, Biesialska, Bojar, Costa-juss{\`a},
  Federmann, Graham, Grundkiewicz, Haddow, Huck, Joanis, Kocmi, Koehn, Lo,
  Ljube{\v{s}}i{\'c}, Monz, Morishita, Nagata, Nakazawa, Pal, Post, and
  Zampieri}]{barrault-etal-2020-findings}
Lo{\"\i}c Barrault, Magdalena Biesialska, Ond{\v{r}}ej Bojar, Marta~R.
  Costa-juss{\`a}, Christian Federmann, Yvette Graham, Roman Grundkiewicz,
  Barry Haddow, Matthias Huck, Eric Joanis, Tom Kocmi, Philipp Koehn, Chi-kiu
  Lo, Nikola Ljube{\v{s}}i{\'c}, Christof Monz, Makoto Morishita, Masaaki
  Nagata, Toshiaki Nakazawa, Santanu Pal, Matt Post, and Marcos Zampieri. 2020.
\newblock \href {https://aclanthology.org/2020.wmt-1.1} {Findings of the 2020
  conference on machine translation ({WMT}20)}.
\newblock In \emph{Proceedings of the Fifth Conference on Machine Translation},
  pages 1--55, Online. Association for Computational Linguistics.

\bibitem[{Edunov et~al.(2018)Edunov, Ott, Auli, and
  Grangier}]{edunov2018understanding}
Sergey Edunov, Myle Ott, Michael Auli, and David Grangier. 2018.
\newblock Understanding back-translation at scale.
\newblock In \emph{Proceedings of the 2018 Conference on Empirical Methods in
  Natural Language Processing}, pages 489--500.

\bibitem[{Enarvi et~al.(2020)Enarvi, Amoia, Teba, Delaney, Diehl, Hahn, Harris,
  McGrath, Pan, Pinto et~al.}]{enarvi2020generating}
Seppo Enarvi, Marilisa Amoia, Miguel Del-Agua Teba, Brian Delaney, Frank Diehl,
  Stefan Hahn, Kristina Harris, Liam McGrath, Yue Pan, Joel Pinto, et~al. 2020.
\newblock Generating medical reports from patient-doctor conversations using
  sequence-to-sequence models.
\newblock In \emph{Proceedings of the first workshop on natural language
  processing for medical conversations}, pages 22--30.

\bibitem[{Gao et~al.(2022)Gao, He, Wu, and Wang}]{gao-etal-2022-bi}
Pengzhi Gao, Zhongjun He, Hua Wu, and Haifeng Wang. 2022.
\newblock \href {https://doi.org/10.18653/v1/2022.naacl-main.289}
  {{B}i-{S}im{C}ut: A simple strategy for boosting neural machine translation}.
\newblock In \emph{Proceedings of the 2022 Conference of the North American
  Chapter of the Association for Computational Linguistics: Human Language
  Technologies}, pages 3938--3948, Seattle, United States. Association for
  Computational Linguistics.

\bibitem[{Kingma and Ba(2014)}]{kingma2014adam}
Diederik~P Kingma and Jimmy Ba. 2014.
\newblock Adam: A method for stochastic optimization.
\newblock \emph{arXiv preprint arXiv:1412.6980}.

\bibitem[{Liu et~al.(2020)Liu, Gu, Goyal, Li, Edunov, Ghazvininejad, Lewis, and
  Zettlemoyer}]{liu2020multilingual}
Yinhan Liu, Jiatao Gu, Naman Goyal, Xian Li, Sergey Edunov, Marjan
  Ghazvininejad, Mike Lewis, and Luke Zettlemoyer. 2020.
\newblock Multilingual denoising pre-training for neural machine translation.
\newblock \emph{Transactions of the Association for Computational Linguistics},
  8:726--742.

\bibitem[{Mota et~al.(2022)Mota, Cabarr{\~a}o, and Farah}]{mota2022fast}
Pedro Mota, Vera Cabarr{\~a}o, and Eduardo Farah. 2022.
\newblock Fast-paced improvements to named entity handling for neural machine
  translation.
\newblock In \emph{Proceedings of the 23rd Annual Conference of the European
  Association for Machine Translation}, pages 141--149.

\bibitem[{Norouzi et~al.(2016)Norouzi, Bengio, Jaitly, Schuster, Wu, Schuurmans
  et~al.}]{norouzi2016reward}
Mohammad Norouzi, Samy Bengio, Navdeep Jaitly, Mike Schuster, Yonghui Wu, Dale
  Schuurmans, et~al. 2016.
\newblock Reward augmented maximum likelihood for neural structured prediction.
\newblock \emph{Advances In Neural Information Processing Systems}, 29.

\bibitem[{Ott et~al.(2019)Ott, Edunov, Baevski, Fan, Gross, Ng, Grangier, and
  Auli}]{ott2019fairseq}
Myle Ott, Sergey Edunov, Alexei Baevski, Angela Fan, Sam Gross, Nathan Ng,
  David Grangier, and Michael Auli. 2019.
\newblock fairseq: A fast, extensible toolkit for sequence modeling.
\newblock In \emph{Proceedings of the 2019 Conference of the North American
  Chapter of the Association for Computational Linguistics (Demonstrations)},
  pages 48--53.

\bibitem[{Post(2018)}]{post2018call}
Matt Post. 2018.
\newblock A call for clarity in reporting bleu scores.
\newblock In \emph{Proceedings of the Third Conference on Machine Translation:
  Research Papers}, pages 186--191.

\bibitem[{Sennrich et~al.(2016{\natexlab{a}})Sennrich, Haddow, and
  Birch}]{sennrich2016improving}
Rico Sennrich, Barry Haddow, and Alexandra Birch. 2016{\natexlab{a}}.
\newblock Improving neural machine translation models with monolingual data.
\newblock In \emph{Proceedings of the 54th Annual Meeting of the Association
  for Computational Linguistics (Volume 1: Long Papers)}, pages 86--96.

\bibitem[{Sennrich et~al.(2016{\natexlab{b}})Sennrich, Haddow, and
  Birch}]{sennrich2016neural}
Rico Sennrich, Barry Haddow, and Alexandra Birch. 2016{\natexlab{b}}.
\newblock Neural machine translation of rare words with subword units.
\newblock In \emph{Proceedings of the 54th Annual Meeting of the Association
  for Computational Linguistics (Volume 1: Long Papers)}, pages 1715--1725.

\bibitem[{Shahhosseini et~al.(2022)Shahhosseini, Hu, and
  Pham}]{shahhosseini2022optimizing}
Mohsen Shahhosseini, Guiping Hu, and Hieu Pham. 2022.
\newblock Optimizing ensemble weights and hyperparameters of machine learning
  models for regression problems.
\newblock \emph{Machine Learning with Applications}, 7:100251.

\bibitem[{Tan et~al.(2022)Tan, Zhang, Wang, and Liu}]{tan2022msp}
Zhixing Tan, Xiangwen Zhang, Shuo Wang, and Yang Liu. 2022.
\newblock Msp: Multi-stage prompting for making pre-trained language models
  better translators.
\newblock In \emph{Proceedings of the 60th Annual Meeting of the Association
  for Computational Linguistics (Volume 1: Long Papers)}, pages 6131--6142.

\bibitem[{Tang et~al.(2020)Tang, Tran, Li, Chen, Goyal, Chaudhary, Gu, and
  Fan}]{tang2020multilingual}
Yuqing Tang, Chau Tran, Xian Li, Peng-Jen Chen, Naman Goyal, Vishrav Chaudhary,
  Jiatao Gu, and Angela Fan. 2020.
\newblock Multilingual translation with extensible multilingual pretraining and
  finetuning.
\newblock \emph{arXiv preprint arXiv:2008.00401}.

\bibitem[{Tran et~al.(2021)Tran, Bhosale, Cross, Koehn, Edunov, and
  Fan}]{tran2021facebook}
Chau Tran, Shruti Bhosale, James Cross, Philipp Koehn, Sergey Edunov, and
  Angela Fan. 2021.
\newblock Facebook ai’s wmt21 news translation task submission.
\newblock In \emph{Proceedings of the Sixth Conference on Machine Translation},
  pages 205--215.

\bibitem[{Tran et~al.(2016)Tran, Dinh, and Nguyen}]{tran2016character}
Phuoc Tran, Dien Dinh, and Hien~T Nguyen. 2016.
\newblock A character level based and word level based approach for
  chinese-vietnamese machine translation.
\newblock \emph{Computational Intelligence and Neuroscience}, 2016:21.

\bibitem[{Van~Nguyen et~al.(2022)Van~Nguyen, Nguyen, Le, Nguyen, Van~Bui, Pham,
  Phan, Nguyen, Tran, and Tran}]{van2022kc4mt}
Vinh Van~Nguyen, Ha~Nguyen, Huong~Thanh Le, Thai~Phuong Nguyen, Tan Van~Bui,
  Luan~Nghia Pham, Anh~Tuan Phan, Cong Hoang-Minh Nguyen, Viet~Hong Tran, and
  Anh~Huu Tran. 2022.
\newblock Kc4mt: A high-quality corpus for multilingual machine translation.
\newblock In \emph{Proceedings of the Thirteenth Language Resources and
  Evaluation Conference}, pages 5494--5502.

\bibitem[{Vaswani et~al.(2017)Vaswani, Shazeer, Parmar, Uszkoreit, Jones,
  Gomez, Kaiser, and Polosukhin}]{vaswani2017attention}
Ashish Vaswani, Noam Shazeer, Niki Parmar, Jakob Uszkoreit, Llion Jones,
  Aidan~N Gomez, {\L}ukasz Kaiser, and Illia Polosukhin. 2017.
\newblock Attention is all you need.
\newblock \emph{Advances in neural information processing systems}, 30.

\bibitem[{Wang et~al.(2021)Wang, Xu, Guzm{\'a}n, El-Kishky, Rubinstein, and
  Cohn}]{wang2021easy}
Jun Wang, Chang Xu, Francisco Guzm{\'a}n, Ahmed El-Kishky, Benjamin Rubinstein,
  and Trevor Cohn. 2021.
\newblock As easy as 1, 2, 3: Behavioural testing of nmt systems for numerical
  translation.
\newblock In \emph{Findings of the Association for Computational Linguistics:
  ACL-IJCNLP 2021}, pages 4711--4717.

\bibitem[{Zhao et~al.(2013)Zhao, Yin, and Zhang}]{zhao2013vietnamese}
Hai Zhao, Tianjiao Yin, and Jingyi Zhang. 2013.
\newblock Vietnamese to chinese machine translation via chinese character as
  pivot.
\newblock In \emph{Proceedings of the 27th Pacific Asia Conference on Language,
  Information, and Computation (PACLIC 27)}, pages 250--259.

\end{thebibliography}
\bibliographystyle{acl_natbib}

% \section*{Acknowledgments}

\appendix
\label{sec:appendix}

\section{Hyper-parameters}
\label{appendix:hyper-params}
We present the detail hyper-parameters used for training our systems in Table~\ref{tab:parameters}.

\begin{table}[t!]
\centering
\begin{tabular}{|l|r|}
\hline
\textbf{Parameter} & \textbf{Value} \\
\hline
Max training updates & 120,000 \\
Early stop patience & 10 \\
Optimizer & Adam \\
Adam eps & 1e-06 \\
Adam $\beta$ &  [0.9, 0.98] \\
Warmup updates & 2,500 \\
Learning rate & 3e-05 \\
Dropout & 0.3 \\
Attention dropout & 0.1 \\
Max tokens & 1,024 \\
Save interval updates & 5,000 \\
\hline
\end{tabular}
\caption{\label{tab:parameters} Hyper-parameters for training our systems}
\end{table}

% \section{Supplemental Material}
% \label{sec:supplemental}

\end{document}